\documentclass[runningheads]{llncs}

\usepackage{eccv}

\usepackage{eccvabbrv}

\usepackage[misc]{ifsym}
\usepackage{pifont}
\usepackage{graphicx}
\usepackage{booktabs}
\usepackage{threeparttable}
\usepackage{lipsum}
\newcommand\blfootnote[1]{%
  \begingroup
  \renewcommand\thefootnote{}\footnote{#1}%
  \addtocounter{footnote}{-1}%
  \endgroup
}

\usepackage{floatrow}
\usepackage[accsupp]{axessibility}  

\usepackage{hyperref}
\usepackage{enumitem}
\usepackage{sidecap}
\sidecaptionvpos{figure}{t}

\usepackage{orcidlink}
\def\eg{\emph{e.g.}} 
\def\ie{\emph{i.e}} 
\newcommand{\Tref}[1]{Table~\ref{#1}}
\newcommand{\Eref}[1]{Eq.~\ref{#1}}
\newcommand{\Fref}[1]{Fig.~\ref{#1}}
\newcommand{\Sref}[1]{Sec.~\ref{#1}}

\usepackage{amsmath,amsfonts,bm}

\def\eqref#1{equation~\ref{#1}}

\def\1{\bm{1}}

\def\rvx{{\mathbf{x}}}

\DeclareMathAlphabet{\mathsfit}{\encodingdefault}{\sfdefault}{m}{sl}
\SetMathAlphabet{\mathsfit}{bold}{\encodingdefault}{\sfdefault}{bx}{n}

\begin{document}

\title{Connecting Consistency Distillation to Score Distillation for Text-to-3D Generation} 

\titlerunning{Connecting Consistency Distillation to Score Distillation}

\author{Zongrui Li \inst{1,2~\dag}\orcidlink{0009-0003-4357-9593} \and
Minghui Hu \inst{2~\dag}\orcidlink{0000-0002-3658-0890}\and
Qian Zheng \inst{3,4} \textsuperscript{\Letter}\orcidlink{0000-0003-3968-3622}\and 
Xudong Jiang \inst{2}\orcidlink{0000-0002-9104-2315} }

\authorrunning{Z. Li et al.}
\institute{ Rapid-Rich Object Search (ROSE) Lab, 
Nanyang Technological University \and
School of Electrical and Electronic Engineering, Nanyang Technological University \and College of Computer Science and Technology, Zhejiang University \and The State Key Lab of Brain-Machine Intelligence, Zhejiang University \\
\email{\{ZONGRUI001, e200008\}@e.ntu.edu.sg, qianzheng@zju.edu.cn, EXDJiang@ntu.edu.sg}
\blfootnote{$^\dag$ Equal contribution. \Letter~Corresponding author.}
}

\maketitle
\begin{abstract}
  Although recent advancements in text-to-3D generation have significantly improved generation quality, issues like limited level of detail and low fidelity still persist, which requires further improvement.
  To understand the essence of those issues, we thoroughly analyze current score distillation methods by connecting theories of consistency distillation to score distillation. Based on the insights acquired through analysis, we propose an optimization framework, Guided Consistency Sampling (GCS), integrated with 3D Gaussian Splatting (3DGS) to alleviate those issues. 
  Additionally, we have observed the persistent oversaturation in the rendered views of generated 3D assets. From experiments, we find that it is caused by unwanted accumulated brightness in 3DGS during optimization.
  To mitigate this issue, we introduce a Brightness-Equalized Generation (BEG) scheme in 3DGS rendering.
  Experimental results demonstrate that our approach generates 3D assets with more details and higher fidelity than state-of-the-art methods.
  The codes are released at \url{https://github.com/LMozart/ECCV2024-GCS-BEG}.
\end{abstract}

\section{Introduction}
\label{sec:intro}

Text-to-3D generation~\cite{liang2023luciddreamer,poole2022dreamfusion,wang2024prolificdreamer,wu2024consistent3d,chen2023fantasia3d,jun2023shap,nichol2022point} has gained substantial attention due to its great potential and indispensable role in many applications, such as gaming, filmmaking, and architecture.
While an end-to-end text-to-3D generative model~\cite{jun2023shap, nichol2022point} is often difficult to train and lacks versatility due to the limited 3D assets in the training set, distillation 3D assets from a well-trained 2D generative model (\eg, diffusion model (DM)~\cite{ho2020denoising, rombach2022high}) has become a popular option due to its data-free feature.
A typical work, DreamFusion~\cite{poole2022dreamfusion}, transfers knowledge from the pre-trained diffusion model to a learnable 3D representation through Score Distillation Sampling (SDS), generating text-aligned 3D assets.

Nevertheless, as highlighted in many subsequent works~\cite{wang2024prolificdreamer, wu2024consistent3d, liang2023luciddreamer, yang2023learn, tang2023dreamgaussian}, SDS suffers from a low level of detail in generation results. Such an issue stems from the poor generalization ability of the distillation method~\cite{wang2024prolificdreamer} and the inherent randomness in the sampling process~\cite{liang2023luciddreamer, wu2024consistent3d}. 
Although work like~\cite{wang2024prolificdreamer} is proposed to enhance the generalization ability, it necessitates fine-tuning the diffusion model during training, dramatically extending the training time.
By contrast, more recent works~\cite{wu2024consistent3d, liang2023luciddreamer} aim to mitigate the randomness in the sampling process by introducing Probability Flow Ordinary Differential Equations (PF-ODEs)~\cite{song2020score}, and speed up the distillation process through 3D Gaussian Splatting (3DGS)~\cite{kerbl20233d}, resulting in cost-efficient, high-quality 3D generation.

{
\begin{figure}[t]
    \centering
    \includegraphics[width=0.9\textwidth]{"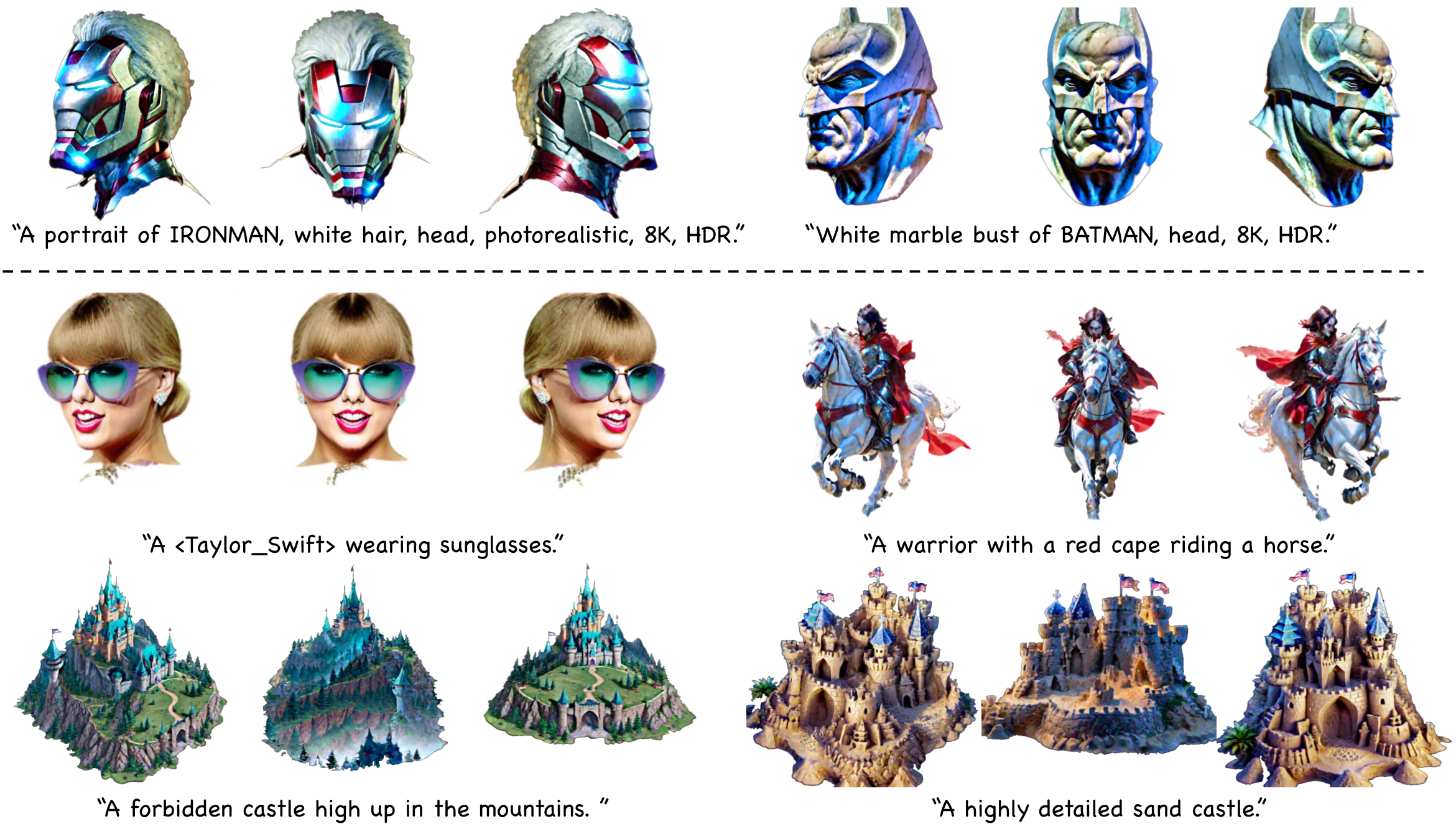"} 
    \caption{Text-to-3D generation results of the proposed \textbf{Guided Consistency Sampling (GCS)} and \textbf{Brightness-equalized Generation}. Each 3D asset is distilled from a pre-trained 2D diffusion model and demonstrated with three different views. The results below the dotted line are generated by the fine-tuned diffusion models.}
    \label{fig:teaser}
\end{figure}
}

Despite the effectiveness of score distillation methods, issues like limited detail and low fidelity remain in the generated 3D asset. 
To reveal the essence of those issues, we conduct an in-depth analysis to connect consistency distillation~\cite{song2023consistency,kim2023consistency, zheng2024trajectory}, a method that distills information from a pre-trained DM through PF-ODEs, to score distillation. Such a novel view helps us identify potential causes of those issues in the score distillation process: 1) the inherent distillation errors of PF-ODEs, 2) ineffective conditional guidance accounting for classifier-free guidance (CFG) effects~\cite{ho2022classifier}, and 3) lack of constraints in pixel domain to avoid out-of-distribution problem, requiring further improvements. 

Except for identified issues in score distillation, there is another noteworthy concern in the 3D rendering process, \ie, the over-saturated views for generated 3D asset~\cite{wang2024prolificdreamer,liang2023luciddreamer,wu2024consistent3d}. While the cause of such an issue is correlated with types of 3D representation, this paper particularly focuses on one (\ie, 3DGS) due to its remarkable reduction in optimization time. Through experiments, we observe that the brightness in generated 2D views accumulates to the next epoch during distillation, eventually leading to over-saturation. 

\begin{figure}[t]
    \centering
    \includegraphics[width=1\textwidth]{"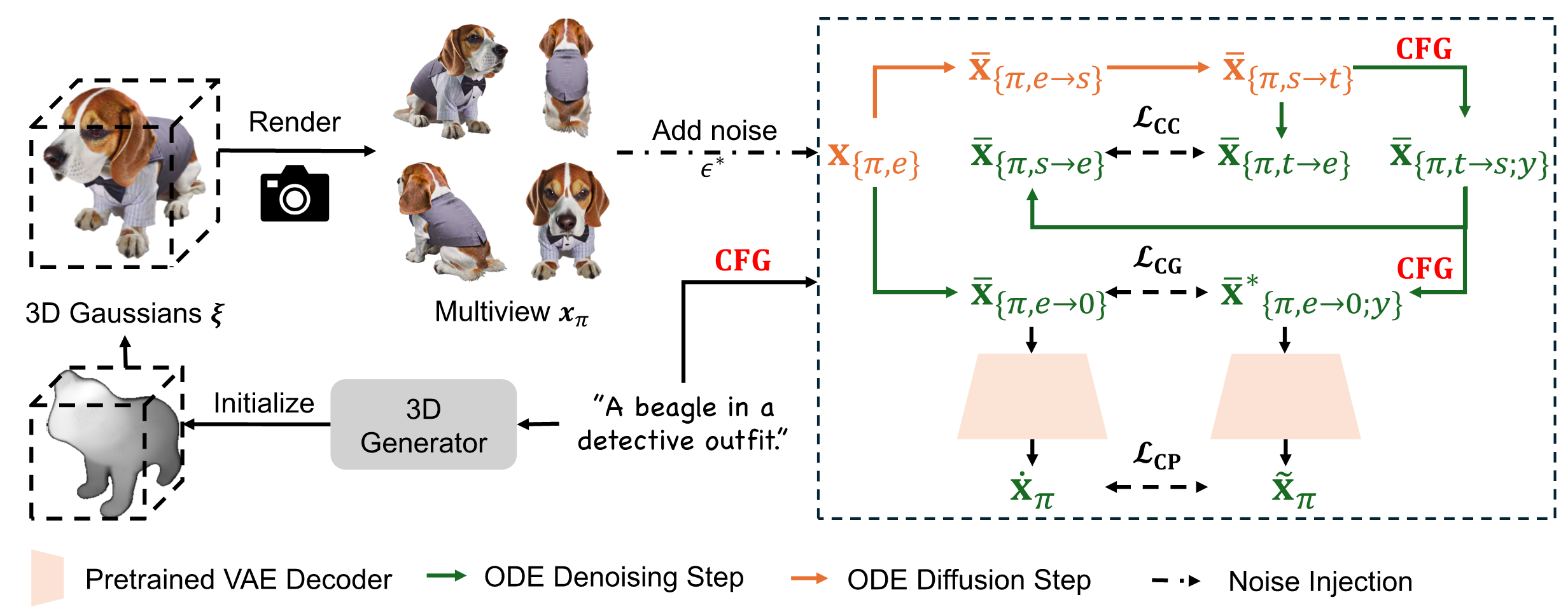"} 
    \caption{An overview of the proposed GCS. We first initialize the 3D representation via the pre-trained 3D generator. For each training epoch, we randomly render a batch of views $\mathbf{x}_\pi$ and diffuse them to $\mathbf{x}_{\{\pi, e\}}$ with a fixed noise $\epsilon^*$. We then apply the ODE diffusion process to gradually add noise to the $\mathbf{x}_{\{\pi, e\}}$ and transfer it to $\bar{\mathbf{x}}_{\{\pi, e\rightarrow s\}}$ and $\bar{\mathbf{x}}_{\{\pi, s\rightarrow t\}}$. In the denoising path, we conduct conditional and unconditional denoising steps, as shown in the figure. Eventually, we calculate the $\mathcal{L}_{\text{GCS}}$ (Eq.~\ref{eq:gcs}) to update the parameters of 3D representation (3DGS). Note that we add `$*$' on  $\bar{\mathbf{x}}_{\{\pi, e \rightarrow 0; y\}}$ to indicate that it is obtained from different sampling trajectories.}
    \label{fig:setup}
\end{figure}

To achieve better performance and overcome the aforementioned issues, this paper proposes \textbf{Guided Consistency Sampling} (GCS) with \textbf{Brightness-equalized Generation} (BEG) for 3DGS-based text-to-3D generation (Fig.~\ref{fig:setup}). The proposed GCS includes three components to solve the identified issues in score distillation: a compact consistency loss reduces distillation errors, a conditional guidance loss provides more effective conditional guidance, and a constraint on pixel domain enhances the fidelity of the 3D assets. 
Additionally, the proposed BEG helps regularize the accumulation of brightness in 3DGS, which significantly alleviates over-saturation issues. 
Examples of generated 3D assets can be found in Fig.~\ref{fig:teaser}.
In summary, our contributions are as follows:

\begin{enumerate}
    \item We identify three problems in the PF-ODEs-based score distillation method by connecting consistency distillation to score distillation.
    \item We propose \textbf{Guided Consistency Sampling}, an improved score distillation method, to enhance the details and fidelity of the generated 3D asset.
    \item We find an accumulated brightness issue in the 3DGS-based rendering process that causes over-saturation and propose \textbf{Brightness-equalized Generation} to alleviate it in the 3DGS-based rendering process.
\end{enumerate}

\section{Related Work}

\textbf{Text-to-3D Generation by Score Distillation.} SDS, introduced in DreamFusion~\cite{poole2022dreamfusion}, and concurrent Score Jacobian Chain (SJC) proposed in~\cite{wang2023score}, have been regarded as milestone works in distilling information from a 2D pre-trained diffusion models to differentiable 3D representations, which inspiring many subsequent Text-to-3D generation methods~\cite{seo2023let, chen2023fantasia3d, lin2023magic3d,chen2024text, wang2024prolificdreamer, zhu2023hifa,chen2024it3d, wu2024consistent3d,tang2023make, tang2023dreamgaussian,metzer2023latent,tsalicoglou2024textmesh,katzir2023noise,shi2023mvdream,zhang2024avatarverse,liu2023zero,zhang2023avatarstudio,yu2023text} to improve the quality of generated 3D assets. Generally, they either endow the DM with the pose-aware generative ability~\cite{metzer2023latent,liu2023zero,zhang2023avatarstudio,shi2023mvdream,zhang2024avatarverse,tsalicoglou2024textmesh} to alleviate the Janus problem~\cite{armandpour2023re} for a more consistent 3D asset, or redesign the distillation process and optimization objectives~\cite{chen2023fantasia3d, lin2023magic3d,zhu2023hifa,wang2024prolificdreamer,yu2023text,katzir2023noise,wu2024consistent3d,chen2024it3d} to reduce the distillation error from the 2D pre-trained diffusion model for a higher level of details. While the performance of the former methods may be restricted by the inherent limitations of SDS~\cite{liang2023luciddreamer}, the latter methods dedicated to overcoming these limitations have recently gained increasing attention. 
For instance, ProlificDreamer~\cite{wang2024prolificdreamer} proposes sampling from the parameter distribution instead of seeking the optimal solution for high-fidelity generation. CSD~\cite{ho2022classifier} introduces insights into the effects of SDS's components and proposes improved strategies. More recent works like~\cite{liang2023luciddreamer,wu2024consistent3d} introduce PF-ODEs into the SDS for a stable, deterministic distillation process. 
After thoroughly investigating recent works, we recognize a connection between consistency distillation and score distillation, which motivates us to analyze and redesign the optimization objectives in view of consistency distillation.
Although concurrent work~\cite{wu2024consistent3d} also highlighted the effectiveness of text-to-3D generation using consistency distillation, our analysis delves deeper into such connection from a broader perspective. We also compare with the findings of~\cite{wu2024consistent3d} in Tab.~\ref{tab:combined}. Another concurrent study~\cite{mcallister2024rethinking} established a relationship between SDS and bridge matching~\cite{chen2016relation,leonard2013survey}, leading to an improvement similar to that seen in our method. 
Guided by acquired insights, we propose GCS that integrates techniques in CM to improve generation quality and gain state-of-the-art performance.

\noindent\textbf{Differentiable 3D Representations}, such as NeRF~\cite{wang2024prolificdreamer,liang2023luciddreamer,poole2022dreamfusion,mildenhall2021nerf}, 3DGS~\cite{kerbl20233d,tang2023dreamgaussian,chen2024text}, or learnable intrinsic that integrated with differentiable geometry~\cite{chen2023fantasia3d,shen2021deep} and textures~\cite{burley2012physically,li2023dani,deng2024flashtex,bensadoun2024meta}, can be optimized by minimizing the loss between rendered and ground truth images, establishing a connection between 3D and 2D representations. Text-to-3D by distillation relies heavily on differentiable 3D representations. While NeRF is well-known to be slow to optimize, we use 3DGS considering its efficiency and comparable performance with NeRF.

\section{Preliminary}
\noindent\textbf{Diffusion Models (DM)}~\cite{ho2020denoising, saharia2022photorealistic} apply a forward process to perturb data samples $\mathbf{x}_0$ drawn from a data distribution $p_{\text{data}}(\mathbf{x})$ by the Gaussian noise $\mathcal{N}(\mathbf{0}, \sigma^2_t\mathbf{I})$ with time-dependent variance $\sigma_t$, where $t \in [0, T]$. As such, the distribution of noisy samples $\mathbf{x}_t$ at time-step $t$ is:
{
\begin{align}
p_t\left(\mathbf{x}_t \mid \mathbf{x}_0\right)=\mathcal{N}\left(\mathbf{x}_t ; \mathbf{x}_0, \sigma_t^2 \mathbf{I}\right),
\end{align}
}which can be re-parameterized to $\mathbf{x}_t=\alpha_t\mathbf{x}_0+\sigma_t \boldsymbol{\epsilon}_t$, where $\epsilon_t \sim \mathcal{N}(\mathbf{0}, \mathbf{I})$.
A reverse process can denoise the noisy samples from $\mathbf{x}_t$ to $\mathbf{x}_0$, expressed as variational inference of Markov processes~\cite{ho2020denoising}.
For all diffusion processes, there exists a corresponding deterministic process with trajectory sharing the same marginal distribution, known as PF-ODE~\cite{song2020score}:
{
\begin{equation}
    \text{\it PF-ODE: }  \  \mathrm{d} \mathbf{x}_t=\left[\mathbf{f}(\mathbf{x}_t, t)-\frac{1}{2} g(t)^2 \nabla_{\mathbf{x}_t} \log p_t(\mathbf{x}_t)\right] \mathrm{d} t,
\end{equation}
}where $\mathbf{f}(\mathbf{x}_t, t)$ is the drift coefficients, $\mathrm{d} t$ is the infinitesimal negative timestep, $\nabla_{\mathbf{x}_t} \log p_t(\mathbf{x}_t)$ is the score function of $p_t(\mathbf{x}_t)$, estimated by a neural network $\boldsymbol{\epsilon}_\theta\left(\mathbf{x}_t, t\right)$. An ODE solver (\eg, DDIM~\cite{song2020denoising}, DPM-Solver~\cite{lu2022dpm}, etc.) can be used to derive the solution of ODE, identical as a sampling process.
The sampling trajectory of ODE is deterministic, and the randomness comes from the initial value, which is more stable than a random sampling process described by the inverse Stochastic Differentiable Equation (SDE)~\cite{song2020score}.

\noindent \textbf{Consistency Model (CM)}~\cite{song2023consistency} is proposed to facilitate a single-step or low number of function evaluations (NFEs)~\cite{pham2015reduction} generation by distilling knowledge from pre-trained DM models. It defines a one-step generator $\boldsymbol{f}_\theta(.; .)$ with trainable parameters $\theta$ that directly predicts the denoised image $\mathbf{x}_0$ given $t$ and $\mathbf{x}_t$, constrained by some boundary conditions~\cite{song2023consistency}. $\boldsymbol{f}_\theta(.; .)$ is trained by minimizing consistency distillation loss~\cite{song2023consistency} defined as:
{
\begin{align} 
\label{eq:cm}
\mathcal{L}_{\mathrm{CD}}\left(\boldsymbol{\theta}, \boldsymbol{\theta}^{-}\right)= \quad \mathbb{E}\left[\omega(t)\left\|\boldsymbol{f}_{\boldsymbol{\theta}}\left(\mathbf{x}_{t_{n+1}}; t_{n+1}\right)-\boldsymbol{f}_{\boldsymbol{\theta}^-}\left(\bar{\mathbf{x}}_{\{t_{n+1} \rightarrow t_n\}}; t_n\right)\right\|_2^2\right],
\end{align}
}where, $0=t_1<t_2 \cdots<t_N=T$, $\bar{\mathbf{x}}_{\{t_{n+1}\rightarrow t_n\}}$ is calculated given ODE solver $\Phi(.)$ as $\bar{\mathbf{x}}_{\{t_{n+1} \rightarrow t_n\}}=\Phi\left(\mathbf{x}_{t_{n+1}}; t_{n+1}, t_n\right)$, $\boldsymbol{\theta}^-$ is updated during training process through an exponential moving average (EMA) strategy~\cite{song2020score}. The ultimate goal of CM is to maintain the self-consistency condition along the trajectory $\{\mathbf{x}_t\}_{t\in [0, T]}$, satisfying,
{
\begin{align}
    \boldsymbol{f}\left(\mathbf{x}_t; t\right)=\boldsymbol{f}\left(\mathbf{x}_{t^{\prime}}; t^{\prime}\right) \quad \forall t, t^{\prime} \in[0, T].
\end{align}
}Subsequent works of CM~\cite{ho2022classifier,zheng2024trajectory, kim2023consistency} integrate Classifier-free guidance (CFG)~\cite{ho2022classifier} or jumps along PF-ODE trajectory into the CM for high-fidelity samples given low or high NFEs, respectively. Inspired by those works, we seek to improve the performance of SDS further in view of CM given their connection.

\section{Methodology}
In \Sref{sec_3_1}, we analyze current advances~\cite{poole2022dreamfusion,yu2023text,wang2024prolificdreamer,liang2023luciddreamer,wu2024consistent3d} in a unified framework obtained by connecting consistency distillation~\cite{kim2023consistency, zheng2024trajectory, song2023consistency} to score distillation. We revisited and identified three problems in the current PF-ODEs-based score distillation from the perspective of consistency distillation. We further proposed our solution, \ie., Guided Consistency Sampling (GCS), and explained in detail in \Sref{sec_4_2}. While GCS is integrated with 3DGS, we observe the accumulated brightness causing over-saturation in 3DGS, as explained in~\Sref{sec:RB}. We propose Brightness Equalized Generation (BEG) to alleviate this issue.

\subsection{Connecting Consistency Distillation to Score Distillation}
\label{sec_3_1}

\begin{table}[t]
  \centering
  \caption{Specific design space employed by proposed GCS and selected 2D diffusion-based score distillation text-to-3D generation method.}
  \resizebox{\textwidth}{!}{
    \begin{tabular}{lccc}
    \toprule
    & \textbf{DreamFusion}~\cite{poole2022dreamfusion} & \textbf{ProlificDreamer}~\cite{wang2024prolificdreamer} & \textbf{Consistent3D}~\cite{wu2024consistent3d} \\
    \midrule
    \textbf{3D Representation} & NeRF / GS & NeRF & NeRF / GS \\
    CFG Weight & - & 7.5+ & 50+ \\
    Objective & SDS (Eq.~\ref{eq:add_eq2}) & VSD (Eq.~\ref{eq:vsd_eq}) & CDS (Eq.~\ref{eq:mse_cd}) \\
    \midrule
    \textbf{Objective \#1: Generator Loss} & & & \\
    term A & $\mathbf{x}_\pi$ & $F_\phi\left(\mathbf{x}_{\{\pi, t\}}; t, y\right)$ & $F_\theta\left(\mathbf{x}_{\{\pi, t\}}; t, y\right)$ \\
    term B & $F_\theta\left(\mathbf{x}_{\{\pi, t\}}; t, y\right)$ & $F_\theta\left(\mathbf{x}_{\{\pi, t\}}; t, y\right)$ & $F_\theta\left(\bar{\mathbf{x}}_{\{\pi, t\rightarrow s; y\}}; s, y\right)$ \\
    \bottomrule
    \toprule
    & \textbf{LucidDreamer}~\cite{liang2023luciddreamer} & \textbf{DreamFusion w/ CFG}~\cite{poole2022dreamfusion} & \textbf{Ours} \\
    \midrule
    \textbf{3D Representation} & NeRF / GS & NeRF / GS & GS w/ BEG \\
    CFG Weight & 7.5+ & 100+ & 7.5+ \\
    Objective & ISD (Eq. ~\ref{eq:isd}) & SDS w/ CFG (Eq.~\ref{eq:cfg_gen})  & GCS (Eqs.~\ref{eq:cc},~\ref{eq:cg},~\ref{eq:cp}) \\
    \midrule
    \textbf{Objective \#1: Generator Loss} & & & \\
    term A & - & $\mathbf{x}_\pi$ & $G_\theta\left(\hat{\mathbf{x}}_{\{\pi, t\}}; t, e, \emptyset\right)$ \\
    term B & - & $F_\theta\left(\mathbf{x}_{\{\pi, t\}}; t, y\right)$ & $F_\theta\left(\bar{\mathbf{x}}_{\{\pi, t \rightarrow s; y\}}; s, e, \emptyset\right)$ \\
    \midrule
    \textbf{Objective \#2: Classifier Loss} & & & \\
    term A & $F_\theta\left(\hat{\mathbf{x}}_{\{\pi, s\}} ; s, \emptyset\right)$  &  $F_\theta\left(\mathbf{x}_{\{\pi, t\}} ; t, \emptyset\right)$ & $F_\theta\left(\hat{\mathbf{x}}_{\{\pi, e\}}; e, \emptyset\right)$ \\
    term B & $F_\theta\left(\hat{\mathbf{x}}_{\{\pi, t\}} ; t, y\right)$   & $F_\theta\left(\mathbf{x}_{\{\pi, t\}} ; t, y\right)$   & $F_\theta\left(G_\theta\left(\hat{\mathbf{x}}_{\{\pi, t\}}; t, e, y\right); e, y\right)$ \\
    \midrule
    \textbf{Pixel Constraint} &  & $+$SDS on pixel domain \cite{zhu2023hifa} & $+\mathcal{L}_{\mathrm{CP}}$ (Eq.~\ref{eq:cp}) \\
    \bottomrule
    \end{tabular}
  }
  \label{tab:combined}
\end{table}

Through theoretical analysis, we find that consistency and score distillation share a similar optimization objective. While CM enforces a {\it self-consistency} of a PF-ODE sampling trajectory (\Eref{eq:cm}), SDS and its variants (score distillation) facilitate a {\it cross-consistency}\footnote{We define {\it cross-consistency} as an alignment between the diffusion trajectory of the rendered view and the text-conditioned sampling trajectory.} between the diffusion trajectory of the rendered view and the denoising trajectory (usually with text condition) of a noisy sample. This cross-consistency is maintained by minimizing the gap between a sample on the diffusion and denoising trajectory. The score distillation method diversifies in choosing those samples to form the optimization objectives (see Tab.~\ref{tab:combined} for a summary). In the following part, we unify the optimization objective of the score distillation method by sample ($\rvx$) instead of score ($\epsilon$) to connect consistency distillation to score distillation based on their similarity.

In vanilla SDS, given a camera poses $\pi$, a 3D representation $\xi$ can be projected to a specific 2D view, noted as $\mathbf{x}_{\pi}=g(\pi, \xi)$. The optimization objective of SDS on 3D parameters $\xi$ is given as:
{
\begin{align}
\label{eq:sds_score}
\min _{\xi} \mathcal{L}_{\mathrm{SDS}}(\xi)=\mathbb{E}_{t, \pi}\left[\omega(t)\left\|\boldsymbol{\epsilon}_{\theta}\left(\mathbf{x}_t; t, y\right)-\boldsymbol{\epsilon}\right\|^2_2\right],
\end{align}
}where $\epsilon_\theta\left(\mathbf{x}_t; t, y\right)$ is the predicted noise given $\mathbf{x}_t$, time-step $t$, and $y$; $\omega(t)$ is a $t$ related weight function. 
As suggested in~\cite{zhou2023sparsefusion, zhu2023hifa,sauer2023adversarial}, SDS loss is equivalent to:
{
\begin{align}
\label{eq:add_eq}
\mathcal{L}_{\mathrm{SDS}}(\xi) & =\mathbb{E}_{t, \pi}\left[\omega(t)\left\|\boldsymbol{\epsilon}_{\theta}\left(\mathbf{x}_{\{\pi, t\}}; t, y\right)-\boldsymbol{\epsilon}\right\|^2_2\right], \\ 
 & = \mathbb{E}_{t, \pi}\left[c(t) \left\| \mathbf{x}_{\{\pi, t\}}- F_\theta(\mathbf{x}_{\{\pi, t\}}; t, y)\right\|_2^2\right],
  \label{eq:add_eq2}\\
 & =: \mathcal{L}_{\text{Distill}}(\xi),
 \label{eq:ft}
\end{align}
}where $\mathbf{x}_{\{\pi, t\}} = \alpha_t \mathbf{x}_\pi + \sigma_t \epsilon$, $F_{\theta}(\mathbf{x}_{\{\pi, t\}}; t, y) = \frac{\mathbf{x}_{\{\pi, t\}}-\sigma_t \boldsymbol{\epsilon}_\theta\left(\mathbf{x}_{\{\pi, t\}}; t, y\right)}{\alpha_t}$, $c(t)$ is another $t$ related weight function. 
In CM~\cite{luo2023latent,ho2020denoising}, $\boldsymbol{f}_\theta(.;.)$ can be parameterize as $F_\theta(.;.)$, which makes Eq.~\ref{eq:cm} similar to Eq.~\ref{eq:add_eq2} in formula. As stated in~\cite{song2020denoising, wu2024consistent3d}, we interpret
minimizing $\mathcal{L}_{\text{SDS}}$ as facilitating a cross-consistency of the stochastic and deterministic trajectory. The variance of SDS injects different conditions on samples. For instance, 
replacing $\mathbf{x}_\pi$ by samples from another fine-tuning diffusion model, noted as $F_\phi(\mathbf{x}_{\{\pi, t\}}; t, y)$ in Eq.~\ref{eq:add_eq2}, we have {\it Variational Score Distillation} (VSD)~\cite{wang2024prolificdreamer} as:
{
\begin{equation}
\label{eq:vsd_eq}
    \mathcal{L}_{\text{VSD}}(\xi) = \mathbb{E}_{t, \pi}\left[c(t) \left\| F_\phi(\mathbf{x}_{\{\pi, t\}}; t, y) - F_\theta(\mathbf{x}_{\{\pi, t\}}; t, y)\right\|_2^2\right],
\end{equation}
}which distills information by enforcing cross-consistency of the trajectory parameterized by $\phi$ and $\theta$.
In practice, CFG with a higher guidance weight $w$ (\eg, $w=100$) is imperative in SDS. Specifically, the expression of SDS with CFG in the form of distillation loss~\cite{ho2022classifier,yu2023text} is:
{
\begin{align}
\label{eq:cfg_gen}
     \mathcal{L}^{\text{CFG}} _{\text{Distill}}(\xi) :=    \mathbb{E}_{t, \pi}[c(t) \Vert & \underbrace{[ \mathbf{x}_\pi-  F_\theta(\mathbf{x}_{\{\pi, t\}}; t, y)]}_\text{generator loss} \nonumber \\ +  & w \underbrace{[F_\theta(\mathbf{x}_{\{\pi, t\}}; t, \emptyset)- F_\theta(\mathbf{x}_{\{\pi, t\}}; t, y)]}_\text{classifier loss} \Vert_2^2].
\end{align}
}Notably, we extend the definition in~\cite{yu2023text}, noting the generator loss as a guidance to make $\rvx_\pi$ more close to the prior distribution and the classifier loss as an update direction for $\rvx_\pi$ to align with the text-condition. As studied in~\cite{yu2023text}, the driving force of SDS is the classifier loss. By omitting the generator loss, they derive {\it Classifier Score Distillation} (CSD), given:
{
\begin{align}
     \mathcal{L}_{\text{CSD}}(\xi) :=    \mathbb{E}_{t, \pi}[c(t) \Vert [F_\theta(\mathbf{x}_{\{\pi, t\}}; t, \emptyset)- F_\theta(\mathbf{x}_{\{\pi, t\}}; t, y)]\Vert_2^2].
\end{align}
}Additionally, as shown in~\cite{liang2023luciddreamer}, $\mathbf{x}_{\{\pi, t\}}$ could be replaced by the deterministic status derived from DDIM to improve the distillation quality:
{
\begin{equation}
\label{eq:deter_x}
    \hat{\mathbf{x}}_{\{\pi, t\}} = \sum^{s}_{k=0} \alpha_{(k+\delta_k)} \left(\frac{1}{\alpha_k} \mathbf{x}_k - \frac{\sigma_{(k+\delta_k)}}{\alpha_{(k+\delta_k)}}\epsilon_\theta(\mathbf{x}_k; k, \emptyset)\right)+ \sigma_{(k+\delta_k)}\epsilon_\theta(\mathbf{x}_k; k, \emptyset),
\end{equation}
}where $t > s$ with $t = s + \delta_{s}$. Similar to derive Eq.~\ref{eq:ft}, based on ~\Eref{eq:deter_x}, we can express {\it Interval Score Distillation} (ISD)~\cite{liang2023luciddreamer} as follows:
{
\begin{equation}
\label{eq:isd}
        \mathcal{L}_{\text{ISD}}(\xi) = \mathbb{E}_{t, s, \pi}\left[c(t) \left\|  F_\theta(\hat{\mathbf{x}}_{\{\pi, s\}}; s, \emptyset) - F_\theta(\hat{\mathbf{x}}_{\{\pi, t\}}; t, y)\right\|_2^2\right],
\end{equation}
}which can be regarded as a classifier loss. With CFG, an additional generator loss is combined with ISD,
{
\begin{align}
    \label{eq:isd_cd}
    \begin{aligned}
        \mathcal{L}^{\text{CFG}}_{\text{ISD}}(\xi)=\mathbb{E}_{t, s, \pi}[c(t)&\|[F_\theta(\hat{\mathbf{x}}_{\{\pi, s\}}; s, \emptyset) - F_\theta(\hat{\mathbf{x}}_{\{\pi, t\}};t, \emptyset)]\\
        &+w[F_\theta(\hat{\mathbf{x}}_{\{\pi, t\}}; t, \emptyset) - F_\theta(\hat{\mathbf{x}}_{\{\pi, t\}}; t, y)],
    \|^2_2],
    \end{aligned}
\end{align}
}we find the generator loss in \Eref{eq:isd_cd} is highly correlated to a {\it Consistency Distillation Sampling} (CDS) studied in~\cite{wu2024consistent3d}:
{
\begin{align}
    \label{eq:mse_cd}
    \mathcal{L}_{\text{CDS}}(\xi)=\mathbb{E}_{t, s, \pi}\left[c(t)\|F_\theta(\mathbf{x}_{\{\pi, t\}}; t, y) - F_\theta(\bar{\mathbf{x}}_{\{\pi, t\rightarrow s; y\}};s, y)\|^2_2\right],
\end{align}
}where, $\bar{\mathbf{x}}_{\{\pi, t\rightarrow s; y\}} =F_\theta\left(\mathbf{x}_{\{\pi, t\}}; t, s, y\right)=\Phi\left(\mathbf{x}_{\{\pi, t\}}; t, s, y\right)$. 
Particularly, CDS is a special case as it mainly guides the generated views to match a particular origin of PF-ODE trajectory from the prior distribution through self-consistency. In such case, an upper bound for the distillation error can be deduced~\cite{wu2024consistent3d}:
{
\begin{align}
\label{eq:error}
\left\| \mathbf{x}_{\pi} - \mathbf{x}_0 \right\|_2=\mathcal{O}\left((\Delta t)^p\right)T.
\end{align}
}where $\mathbf{x}_0\sim p_{\text{data}}(\mathbf{x})$ is an real image, $\Delta t =\max \left\{\left|\delta_{k}\right|\right\}$, $k \in [0, ..., s]$ and $t = s + \delta_{s}$. With CFG, CDS can further facilitate a cross-consistency of text and null condition trajectories. However, it performs poorly with low CFG weight~\cite{wu2024consistent3d}, which needs further improvement. Inspired by the similarity of CM and the optimization objectives of the score distillation method, we are motivated to reformulate the optimization objective in SDS by extending the theories in CM to improve text-to-3D generation quality. 
We sum up our solution in three aspects: 1) we reduce the error bound suggested~\Eref{eq:error} to improve the distillation quality; 2) we provide more reliable guidance during distillation accounting for CFG effects; 3) we implement constraints on pixel domain to enhance the fidelity.

\subsection{Guided Consistency Sampling}
\label{sec_4_2}
In this section, we introduce the Guided Consistency Sampling (GCS), which includes three parts of objectives: a Compact Consistency (CC) loss, a Conditional Guidance (CG) score, and a Constraint on Pixel domain (CP):
{
\begin{equation}
\label{eq:gcs}
    \mathcal{L}_{\text{GCS}}(\xi) = \mathcal{L}_{\text{CC}}(\xi) + \mathcal{L}_{\text{CG}}(\xi) + \mathcal{L}_{\text{CP}}(\xi).
\end{equation}
}

\begin{figure}[t]
\floatbox[{\capbeside\thisfloatsetup{capbesideposition={right,top}}}]{figure}[\FBwidth]
{\caption{Generated views by using $\mathcal{L}_{\text{CC}}$ with different CFG strategies at a low CFG weight ($w=7.5$). While $\mathcal{L}_{\text{CC}}$ (left) implements CFG in only one denoising step, $\mathcal{L}^*_{\text{CC}}$ (right) applies CFG in every ODE denoising step.}\label{fig:cc}}
{\includegraphics[width=4cm]{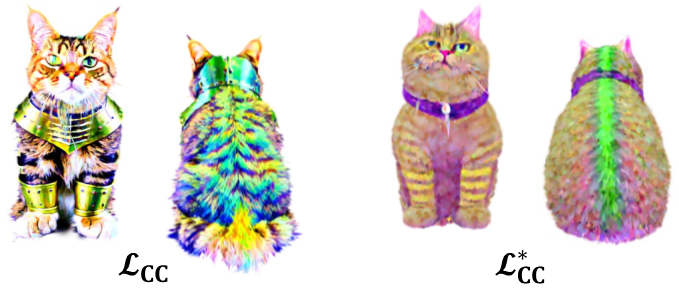}}
\end{figure}

\noindent\textbf{Compact Consistency (CC) Loss} aims to improve further the self-consistency of PF-ODE denoising trajectory, which eventually reduces the distillation error bound in \Eref{eq:error} for a more aligned distribution of rendered views. 
Inspired by~\cite{kim2023consistency, zheng2024trajectory}, we define a solution function:
{
\begin{equation}
    G_{\theta}(\rvx_t; t, s, y) := \rvx_t  + \int_t^s \frac{\rvx_u-\mathbb{E}[\rvx | \rvx_u]}{u} \text{d}u.
\end{equation}
}which $G_{\theta}(\rvx_t; t, s, y)$ solves the PF-ODE from initial time $t$ to a final time $s$ according to exponential integrator~\cite{lu2022dpm, lu2022dpm++, zhang2022fast}. Owing to the fact that $G$ is intractable as it can only be obtained by calculating $s$ partial derivative at time $t$, 
we follow the scheme in~\cite{zheng2024trajectory} and adhere to the first-order definition\footnote{For the effect of higher orders, please refer to the Appendix for more details.} of DPM-Solver~\cite{lu2022dpm}, re-parameterised $G$ as:
{
\begin{equation}
\label{eq:gp}
    G_{\theta}(\rvx_t; t, s, y) = \frac{\sigma_s}{\sigma_t} \rvx_t - \alpha_s (e^{-h} - 1) {\epsilon}_{\bm{\theta}} (\rvx_t, t, y),
\end{equation}
}where $h = \lambda_s - \lambda_t$ with the $\log$-SNR $\lambda$ defined as $\lambda = \log(\alpha / \sigma) $ and ${\epsilon}_{\bm{\theta}}$ is the prediction from the network.
According to the previous definition, we mathematically define Compact Consistency Loss, a critical part of GCS, to improve the details of the 3D asset:
{
\begin{equation}
\label{eq:cc}
    \mathcal{L}_{\text{CC}}(\xi)=\mathbb{E}_{t,s,e, \pi}\left[\|G_\theta(\hat{\mathbf{x}}_{\{\pi, t\}}; t, e, \emptyset) - G_\theta(\bar{\mathbf{x}}_{\{\pi, t \rightarrow s; y\}};s, e, \emptyset)\|^2_2\right],
\end{equation}
}where $t>s>e$,
and $\hat{\mathbf{x}}_{\{\pi, t\}}$ is obtained by DDIM inversion from ${\rvx}_{\{\pi, e\}}$, calculated as ${\rvx}_{\{\pi, e\}} = \alpha_e \rvx_\pi + \sigma_e \epsilon^*$ to $\bar{\rvx}_{\{\pi, e\rightarrow s\}}$ and eventually to $\bar{\rvx}_{\{\pi, s\rightarrow t\}}$.  A null condition is applied to all DDIM inversion steps. Notably, $\epsilon^*$ is the random noise that will only be sampled once and kept fixed in practice\footnote{We use $\epsilon^*$ to ensure a valid diffusion process at low time step and reduce the computational cost, compared to~\cite{liang2023luciddreamer}.}.
During the training process, the parameters in the pre-trained model are frozen.
In this condition, we substantiated Lemma~\ref{the:upcm}, which extends the premises established in~\cite{wu2024consistent3d, song2023consistency}. 
\begin{lemma}[\cite{zheng2024trajectory, song2023consistency, kim2023consistency, wu2024consistent3d}]
\label{the:upcm}
Let $\Delta t =\max \left\{\left|\delta_{k}\right|\right\}$, $k \in [0, ..., n_s]$, where $n_s$ is the index of $\delta$ at time step $s$, and $F_{\theta}(\cdot, \cdot)$ is the origin prediction function grounded on the empirical PF-ODE. Assume $F_{\theta}$ satisfies the Lipschitz condition, if there is a $\rvx_\pi$ satisfying $\mathcal{L}_{\rm{CC}}(\xi) = 0$, given an image $\rvx_0 \sim p_{\text{data}}(\rvx)$, for any $t, s, e\in [\![0,..., T]\!]$ with $t > s > e$,  we have:
{
\begin{equation}
\begin{split}
\mathop{\textnormal{sup}}\limits_{t, e, \rvx_\pi} \Vert \hat{\rvx}_{\{\pi, e\}}, \hat{\rvx}_{\{0,e\}} \Vert_2  = \mathcal{O} \left((\Delta t)^{p}\right) (T - e),
\end{split}
\end{equation}
}$\hat{\rvx}_{\{0,e\}}$ is the distribution of $\rvx_0$ diffused to time $e$, $p$ is the order of ODE solver.
\begin{proof}
The proof is provided in the Appendix for completeness.
\end{proof}
\end{lemma}
Lemma.~\ref{the:upcm} validates that CC achieves a lower upper limit on error margins than CDS (Eq.~\ref{eq:error}) for a better distillation from the pre-trained model. 
Considering the importance of CFG in generating high-quality contents, another issue worth noting is whether it is practical to integrate CFG (or Perp-Neg~\cite{armandpour2023re}) into a generator loss $\mathcal{L}_{\text{CC}}$. In CDS, it applies CFG in every ODE denoising step, which leads to a large accumulated error magnified by CFG~\cite{luo2023latent,mokady2022null}. Instead, we only implement CFG in one step~\cite{luo2023latent} (see \Fref{fig:setup}), leading to lower accumulated errors. This strategy significantly improves the generation quality with low CFG weight (see \Fref{fig:cc}). However, we still observe artifacts in the generated 3D model, likely due to a lack of effective classifier loss that provides text-conditional guidance.
\subsubsection{Conditional Guidance (CG) Loss} is proposed to accommodate guidance while maintaining a lower accumulated error, which we interpret as conditional guidance.  
Inspired by~\cite{liang2023luciddreamer,mokady2022null, ho2022classifier}, we facilitate an alignment between the unconditional trajectory ($\mathcal{T}_{\{e\rightarrow 0, \emptyset\}}$ from a noisy sample at $e$ to $0$) and the text-conditional trajectory ($\mathcal{T}_{\{t\rightarrow e \rightarrow 0, y\}}$ derived from a noisy sample $\hat{\rvx}_{\{\pi, t\}}$ then sampling down to $e$ and $0$), separately. Mathematically, the CG score can be expressed as:
{
\begin{equation}
\label{eq:cg}
    \mathcal{L}_{\text{CG}}(\xi)=\mathbb{E}_{t, e, \pi}\left[\| F_\theta(\hat{\mathbf{x}}_{\{\pi, e \}}; e, \emptyset) - F_\theta(G_\theta(\hat{\mathbf{x}}_{\{\pi, t\}};t, e, y); e, y)\|^2_2\right],
\end{equation}
}where $F_\theta(\rvx; t, y)$ predicts the $\rvx_0$ from time step $t$ given the conditional information $y$ as shown in Eq.~\ref{eq:ft}, and $G_\theta(\rvx; t, e, y)$ access the midpoint $e$ of the trajectory from initial timestep $t$ as explained in Eq.~\ref{eq:gp}. In practice, we can integrate CFG into $\mathcal{L}_{\text{CG}}$ for better performance. We believe the proposed $\mathcal{L}_{\text{CG}}$ provides more reliable guidance with less accumulated error, as it avoids long navigation along the trajectory. Specifically, the gap between $\mathcal{T}_{\{e \rightarrow 0, \emptyset\}}$ and $\mathcal{T}_{\{t\rightarrow e\rightarrow 0, y\}}$ is a more precise guidance 
since it directly affects $\hat{\rvx}_{\{\pi, e\}}$, which is a more similar sample to $\rvx_{\pi}$~\cite{mokady2022null} due to the DDIM inversion.

It is also interesting to see that $\mathcal{L}_{\text{ISD}}$ is equivalent to $\mathcal{L}_{\text{CG}}$ when set $e=0$ in computing $F_\theta(G_\theta(\hat{\mathbf{x}}_{\{\pi, t\}};t, e, y); e, y)$. 
We intuitively explain the effect of a midpoint $e$ to reduce the error of one-step denoising from a noisy sample conditioned at large $t$. Eventually, we find $\mathcal{L}_{\text{CC}}$ and $\mathcal{L}_{\text{CG}}$ work together improving the quality of generated 3D asset.

\subsubsection{Constraint on Pixel (CP) Domain} aims to achieve a closer resemblance in the pixel domain for $\mathcal{L}_{\text{CG}}$. While a satisfactory $\rvx_\pi$ is obtained in the latent domain, whether such an $\rvx_\pi$ is equally satisfactory in the pixel domain appears to have been overlooked. In fact, the optimized $\rvx_\pi$ may not adhere to the prior distribution stipulations of VAE, leading to out-of-distribution artifacts. Consequently, this leads to unrealistic color in the decoded images (\eg, a paint-like color in the generated results of LucidDreamer~\cite{liang2023luciddreamer}). Inspired by similar methods proposed in~\cite{zhu2023hifa}, we further calculate $\mathcal{L}_{\text{CG}}$ in the pixel domain for enhanced supervision. In this context, given an image decoder $\mathcal{D}$ along with $\hat{\mathbf{x}}_{\{\pi, e \}}$, the definition of CP can be articulated as follows:
{
\begin{align}
\label{eq:cp}
    &\mathcal{L}_{\text{CP}}(\xi)=\mathbb{E}_{t, e, \pi}\left[\| \mathcal{D}(\dot{\rvx}_\pi) - \mathcal{D}(\tilde{\rvx}_\pi)\|^2_2\right],\\
    & \dot{\rvx}_\pi = F_\theta(G_\theta(\hat{\mathbf{x}}_{\{\pi, t\}}; t, e, y); e, y),  \\
    & \tilde{\rvx}_\pi =  F_\theta(\hat{\mathbf{x}}_{\{\pi, e \}}; e, \emptyset).
\end{align}
}Our empirical observations indicate that optimizing CP yields notable enhancements in color fidelity, albeit with increased computational cost and over-saturation. Consequently, we suggest an additional strategy outlined in Sec.~\ref{sec:RB}, aimed at alleviating over-saturation. Extensive experiments lead us to conclude that the CP term is constructive in bolstering the overall quality of 3D assets.

\subsection{Brightness-equalized Generation}
\label{sec:RB}
A common problem in previous works is that the generated 3D assets often suffer from over-saturation, especially in those that use Gaussian Splatting as the 3D representation to be trained. 
We experimentally find that the high brightness region (highlight) generated in the current training epoch is carried over to the next epoch\footnote{A visualization of the brightness accumulation can be found in the Appendix.}.
That is, the brightness of the highlight points accumulates during training, eventually leading to over-saturation. To alleviate this issue, we propose resetting the brightness of the Gaussian\footnote{We use the average of RGB channels as the brightness.} adaptively according to the exposure status of the generated view. Specifically, in each training epoch, we calculate the $m^{\mathrm{th}}$(85) percentile of each image $\mathbf{x}^i_\pi$ in an image batch $\mathbf{X}_\pi=\{\mathbf{x}^i_\pi, i\in[1, ..., B]\}$, noted as $P_m = \{p^i_m, i\in[1, ..., B]\}$, where $B$ is the batch size.  We then find the maximum value in $P_m$ and reset the brightness of Gaussian to $T_{\text{B}}=0.8$ of the current brightness if $\operatorname{max}(P_m) > T_{\text{GS}}$, $T_{\text{GS}}=0.9$. Such a simple method will significantly alleviate the over-saturation issues and facilitate brightness equalization.

\section{Experiments}
\subsection{Implementation Details}
We implement the GCS with PyTorch and train it on an A800 GPU with the Adam optimizer. Some hyperparameters (the learning rates, camera positions, rendering resolution, \etc.) are similar to that in~\cite {liang2023luciddreamer}. We use 5000 epochs as~\cite{wang2024prolificdreamer} for total training iterations, which takes about one hour per scene.

\noindent \textbf{Time schedule.} In the experiment, $t$ is sampled from a uniform distribution $\mathcal{U}(20, 500 + \delta_{\text{warm}})$, where $\delta_{\text{warm}}$ is linearly decreasing from $480$ to $0$ in the first 1500 epochs. $s = t - \delta$, where $\delta=100$ for most cases. $e$ is sampled from a uniform distribution as $\mathcal{U}(s- \delta, s - \frac{\delta}{10})$.

\noindent \textbf{Initialization.} We apply Shap-E~\cite{jun2023shap} and Point-E~\cite{nichol2022point} to initialize the 3D Gaussians, which are widely used in 3DGS-based text-to-3D generation~\cite{liang2023luciddreamer, wu2024consistent3d}.
\begin{figure}[t]
    \centering
    \includegraphics[width=\textwidth]{"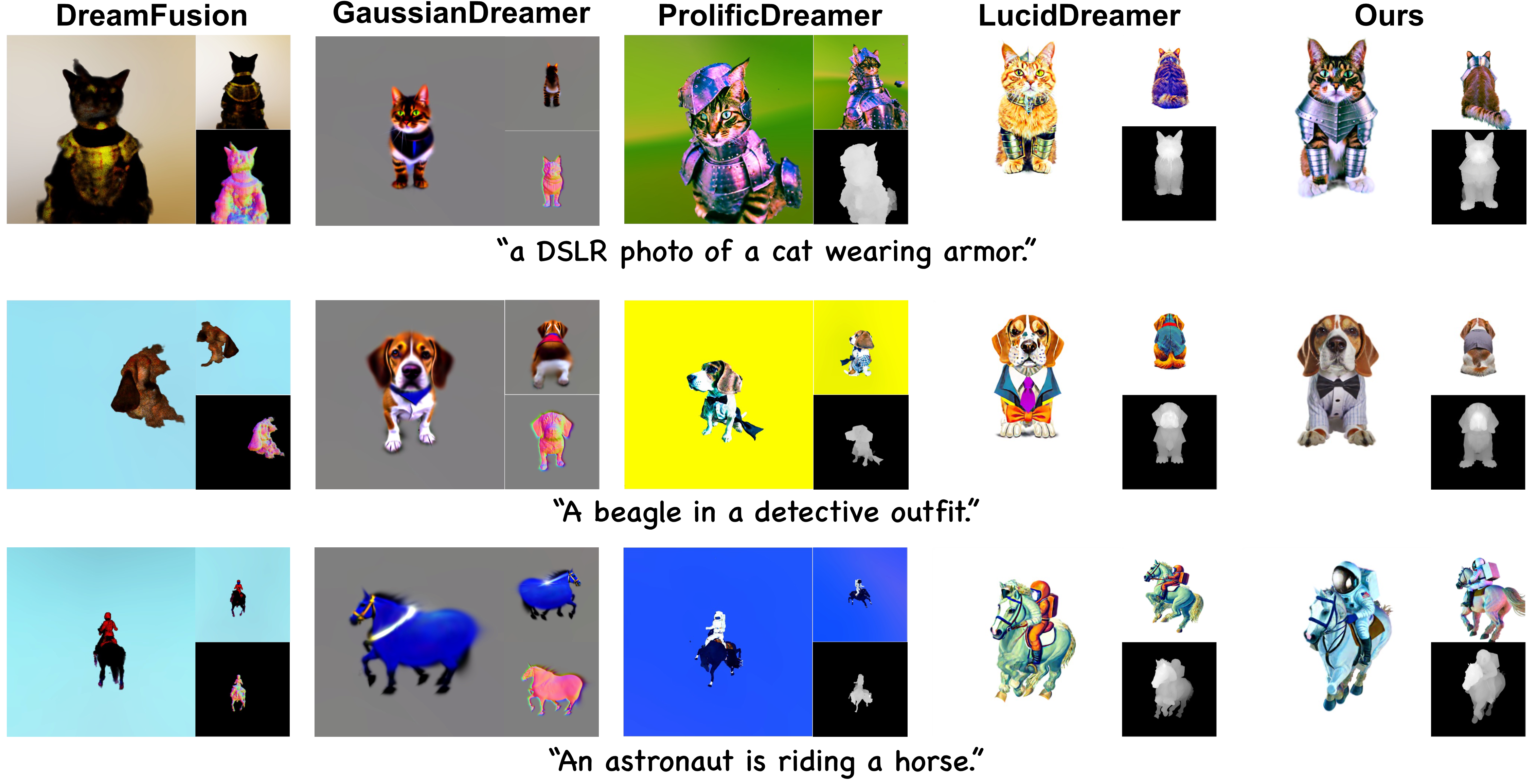"} 
    \caption{Qualitative comparison among the proposed and other methods on text-to-3D generation results. From left to right, results generated by DreamFusion~\cite{poole2022dreamfusion}, GaussianDreamer~\cite{yi2023gaussiandreamer}, ProlificDreamer~\cite{wang2024prolificdreamer}, LucidDreamer~\cite{liang2023luciddreamer}, and the proposed method, with a CFG weight $100$, $100$, $7$, $7$, $7$, respectively. For each sub-figure, left: main view, right-top: back view, right-bottom: normal/depth map of the 3D asset.}
    \label{fig:main_exp}
\end{figure}
\subsection{Text-to-3D Generation} 

\textbf{Qualitative Comparison.} We have showcased diverse views of the 3D assets generated by our versatile approach,
indicating that we can not only generate highly detailed figures but also create reasonable outputs to align with the complex prompt (results shown in the first row). To comprehensively assess the efficacy of our proposed Guided Conditional Sampling (GCS), we conducted a comparative analysis against the current state-of-the-art in GS-based~\cite{liang2023luciddreamer, yi2023gaussiandreamer} and baseline NeRF-based~\cite{poole2022dreamfusion,wang2024prolificdreamer} methods in text-to-3D generation, as shown in \Fref{fig:main_exp}. 
To ensure a fair comparison, all listed methods implement Stable Diffusion 2.1 as the base model. Other hyperparameters were configured following the default settings of the respective methods.
The results indicate that previous methods often struggled with insufficient details, prominent artifacts (\eg, Janus problem~\cite{armandpour2023re}), or over-saturation. In contrast, our approach exhibits notable improvements in detail (\eg, the armor of the cat), fidelity, and exposure. Specifically, we can generate highly detailed textures, achieve photorealistic effects (such as lights on the cat's armor), and maintain more consistent colors (such as the cat’s fur color) even with a low CFG weight, illustrating the effectiveness of our proposed GCS and the brightness-equalization techniques employed during training. Additional results can be found in the Appendix.


\noindent \textbf{Quantitative Comparison.} We present the average CLIP Score, FID Score, and user preference of five comparative methods~\cite{poole2022dreamfusion, wang2024prolificdreamer, liang2023luciddreamer, tang2023dreamgaussian, yi2023gaussiandreamer} and the proposed GCS on 120 generated views across 30 prompts selected from the DreamFusion gallery\footnote{\url{https://dreamfusion3d.github.io/gallery.html}} in \Tref{tab:userstudy}. 
For CLIP Score computation, we follow the steps presented in~\cite{poole2022dreamfusion}. 
Results show that our performance is comparable to that of VSD\cite{wang2024prolificdreamer} in the evaluation of prompt alignment, with much fewer instances of Janus problems~\cite{armandpour2023re}, as evidenced in the qualitative comparison shown in Fig.~\ref{fig:main_exp}.
The proposed approach exhibits improved 3D consistency and fidelity for the 3D asset, resulting in a higher user preference in prompt alignment. 

For FID, we follow similar steps in~\cite{wang2024prolificdreamer} to calculate FID between 120 views of 3D object\footnote{Views are generated with fixed elevation ($0^\circ$, 3.6k views generated in total) and varying azimuth (uniformly covering $360^\circ$)} and 50k Stable Diffusion 2.1 generated images. We implement Perp-Neg when necessary for different objects. The results show that the generated views of the proposed method are more similar to the prior distribution compared to other methods, indicating that the proposed GCS reduces distillation error. 

\noindent \textbf{User Study.} We further conduct a user study among 30 volunteers for a more comprehensive evaluation. Specifically, we asked the volunteers to evaluate the generation quality regarding \textit{brightness} (Q1), \textit{prompt alignment} (Q2), and \textit{fidelity} (Q3). Those volunteers are also required to indicate their preference based on rendered $360^\circ$ videos of 5 objects randomly sampled from 30 examples within six methods. We shuffled the presented order of the generated results to avoid any form of leakage. The results of user preference in percentage form shown in \Tref{tab:userstudy} indicate that our method performs best in all aspects. An example question can be found in the Appendix.

\begin{table*}[t]
  \centering
  \caption{Quantitative comparison regarding CLIP Score, FID Score, and user preference on three given questions (\textbf{Q1}, \textbf{Q2}, and \textbf{Q3}). \textbf{The bold} (\underline{underline}) number indicates the best (second-best) results.}
    \label{tab:userstudy}%
  \resizebox{1\linewidth}{!}{
  \begin{threeparttable}
    \begin{tabular}{lcccccc}
    \toprule
          & \textbf{GaussianDreamer} & \textbf{DreamGaussian} & \textbf{DreamFusion} & \textbf{LucidDreamer} & \textbf{ProlificDreamer} & \textbf{Ours} \\
    \midrule
        \textbf{CLIP Score$\uparrow$} & 30.29 & 28.65 & 30.68  & 31.30  & \textbf{33.17}  & \underline{32.37} \\
    \textbf{FID Score$\downarrow$} & 133.48 & 215.89 & 137.22 & \underline{109.57} & 119.53 & \textbf{103.40} \\
    \textbf{User Preference: Q1$\uparrow$}$^1$ & 9.47  & 1.05  & 6.32  & \underline{24.21} & 12.63 & \textbf{46.32} \\
    \textbf{User Preference: Q2$\uparrow$}$^2$ & 10.53 & 1.05  & 6.32  & \underline{30.53} & 7.37  & \textbf{44.21} \\
    \textbf{User Preference: Q3$\uparrow$}$^3$ & 9.47  & 2.11  & 6.32  & \underline{26.32} & 11.58 & \textbf{44.21} \\
    \bottomrule
    \end{tabular}%
   \begin{tablenotes}    
        \footnotesize    
        \item[1] \textbf{Q1}: From the perspective of \textbf{color brightness}, which of the following methods produces the most balanced brightness (\eg., no over-saturation, moderate saturation)?
        \item[2] \textbf{Q2}: From the perspective of \textbf{prompt alignment}, which of the following methods align most with the prompt?
        \item[3] \textbf{Q3}: From the perspective of \textbf{fidelity}, which of the following methods produces the most realistic objects?
        
      \end{tablenotes}
  \end{threeparttable}
  }
\end{table*}%

\begin{figure}[t]
    \centering
    \includegraphics[width=1.\textwidth]{"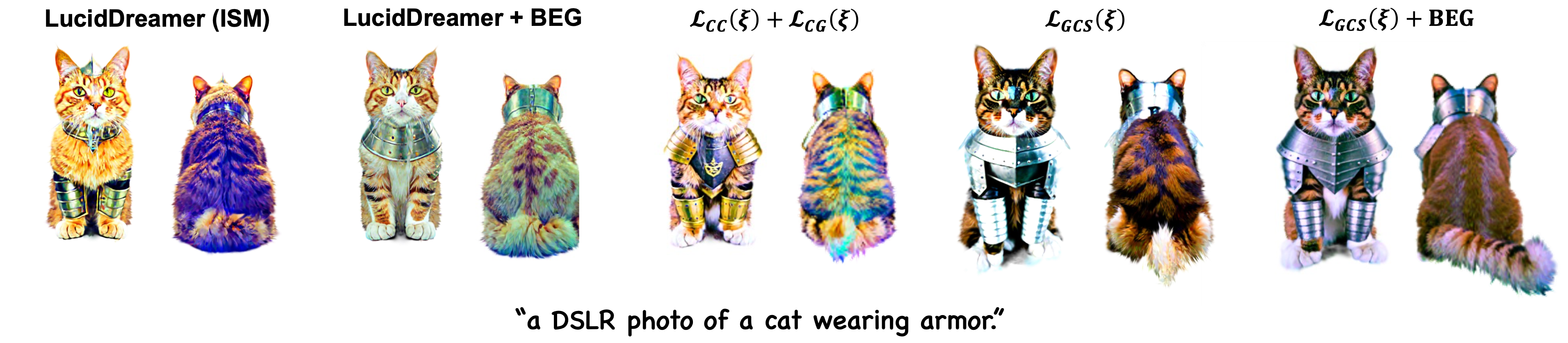"} 
    \caption{Ablation study of proposed components.LucidDreamer~\cite{liang2023luciddreamer} serves as the baseline, we demonstrate the results under settings: (a) LucidDreamer~\cite{liang2023luciddreamer} + BEG, (b) $\mathcal{L}_\mathrm{C C}(\xi)+\mathcal{L}_\mathrm{C G}(\xi)$, (c) $\mathcal{L}_\mathrm{GCS}(\xi)$, and (d) full mode ($\mathcal{L}_\mathrm{GCS}(\xi)$+ BEG) from left to right. }
    \label{fig:abl}
\end{figure}

{
\begin{figure}[t]
    \centering
    \includegraphics[width=\textwidth]{"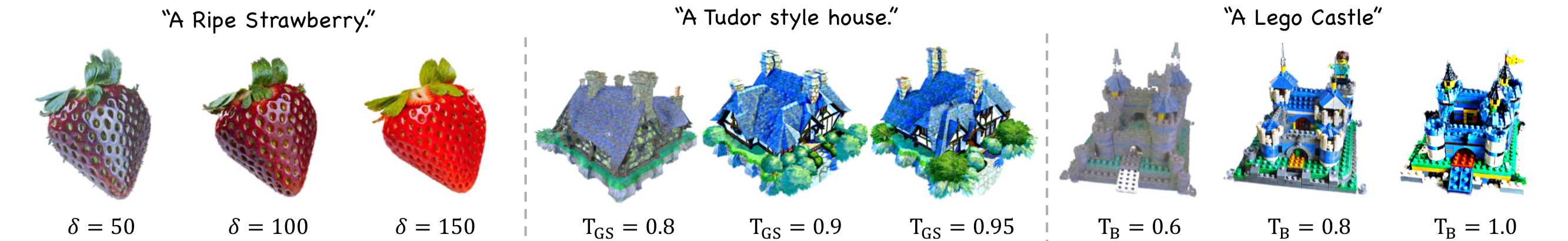"} 
    \caption{Ablation study of hyperparameters' effects (left: $\delta$, middle: $T_{\mathrm{GS}}$, right: $T_{\mathrm{B}}$)}
    \label{fig:abl2}
\end{figure}
}


\subsection{Ablation Study}

We conducted an ablation study to assess the effects of parameter setup (including $\delta$, $T_{\mathrm{GS}}$, and $T_{\mathrm{B}}$), the proposed Guided Consistency Sampling (GCS) loss, and the brightness-equalized generation method (BEG). 



\noindent \textbf{Effect of time interval $\delta$} is evaluated 
by an ablation study on $\delta$ shown in \Fref{fig:abl2}. We find that large $\delta$ causes over-smoothness (\eg, low level of detail on the strawberry's surface), while small $\delta$ causes color distortion (\eg, unrealistic color in strawberry). $100$ is set to generate visual-pleasant results.

\noindent\textbf{Effect of Different Loss and BEG}\footnote{We conduct additional ablation studies regarding CC and CG to analyze the individual effect on generated 3D assets in the Appendix.}. As illustrated in \Fref{fig:abl}, we compared the full model's ($\mathcal{L}_\mathrm{GCS}(\xi)$ + BEG) generated results with those from LucidDreamer~\cite{liang2023luciddreamer}, and two alternative models: $\mathcal{L}_\mathrm{C C}(\xi)+\mathcal{L}_{C G}(\xi)$ and $\mathcal{L}_\mathrm{GCS}(\xi)$ without BEG method, with a CFG weight set to $7.5$. Compared to LucidDreamer~\cite{liang2023luciddreamer}, the full model successfully generated more intricate details (\eg, armor's patterns and higher armor coverage). While there was a slight alleviation in color distortion (inconsistency of fur color between the back and front of the cat), the issue may persist. Consequently, we introduced $\mathcal{L}_{\text{CP}}$ to mitigate color distortion further (comparing $\mathcal{L}_\mathrm{C C}(\xi)+\mathcal{L}_\mathrm{C G}(\xi)$ and $\mathcal{L}_\mathrm{GCS}(\xi)$). 

However, we observed a significant increase in overall brightness during optimization. The implementation of the BEG method notably alleviates this issue, as evidenced in the qualitative comparison between $\mathcal{L}_\mathrm{GCS}(\xi)$ and the full model in~\Fref{fig:abl}. We find that BEG will only affect the overall brightness, but not the fidelity, as shown in a qualitative comparison between LucidDreamer~\cite{liang2023luciddreamer} and LucidDreamer~\cite{liang2023luciddreamer} + BEG. 
The effect of BEG is controlled by $T_{\text{B}}$ and $T_{\text{GS}}$. Specifically, large (small) $T_{\text{B}}$ causes over-saturation (color distortion) results (see the comparison in~\Fref{fig:abl2}). $T_{\text{B}}=0.8$ works for general cases. Small $T_{\text{GS}}$ will reset the brightness too frequently to cause artifacts, $T_{\text{GS}}=0.9$ is a suitable setup.

\section{Conclusion}
In this work, we connect consistency distillation to score distillation. From this connection, we propose Guided Consistency Sampling (GCS), an optimization framework that includes three parts: Compact Consistency (CC) loss for improved generator loss, Conditional Guidance (CG) score to enhance conditional guidance, and Constraints in the Pixel domain (CP) for improved fidelity. In addition, we innovate the Brightness-equalized Generation (BEG) to tackle the over-saturation issue in 3DGS training. The proposed approach improves the details, quality, and lighting effects of the generated 3D assets.

\section*{Acknowledgement} This work is supported by Rapid-Rich Object Search (ROSE) Lab, Nanyang Technological University, Singapore, and the State Key Lab of Brain-Machine Intelligence, Zhejiang University, Hangzhou, China.

\bibliographystyle{splncs04}
\bibliography{main}
\newpage
\appendix
\section*{Appendix}
In this appendix, 
\begin{enumerate}[itemsep=0.pt,topsep=0.pt,parsep=0.pt]
    \item we provide some discussions about the limitations of our method and the failure cases;
    \item we show an example question for the user study in \Sref{sec:example};
    \item we visualize the accumulated brightness during optimization in \Sref{sec:acc};
    \item we provide a detailed proof of Lemma \textcolor{red}{1} in \Sref{sec:proof} that clarifies the effectiveness of the proposed Compact Consistency (CC) loss;
    \item we show more ablation studies regarding CC and Conditional Guidance (CG) loss in \Sref{sec:abl}; we also show the effect of high order ODE solver in this section;
    \item we showcase more results generated by the proposed Guided Consistency Sampling (GCS) in \Sref{sec:qe}.
    
\end{enumerate}

\section{Limitations} The proposed GCS exhibits shortcomings in addressing Janus issues, which we attribute to the inherent inconsistencies in pre-trained models under multi-view conditions. Also, it doesn't perform well on compositional prompts. Moreover, our approach necessitates extensive training in the GS model, ranging from 3000 to 5000 epochs, translating to 35-60 minutes for rendering a single object. These limitations show a significant gap in practical application. 

\begin{figure}[H]
    \centering
    \includegraphics[width=0.9\textwidth]{"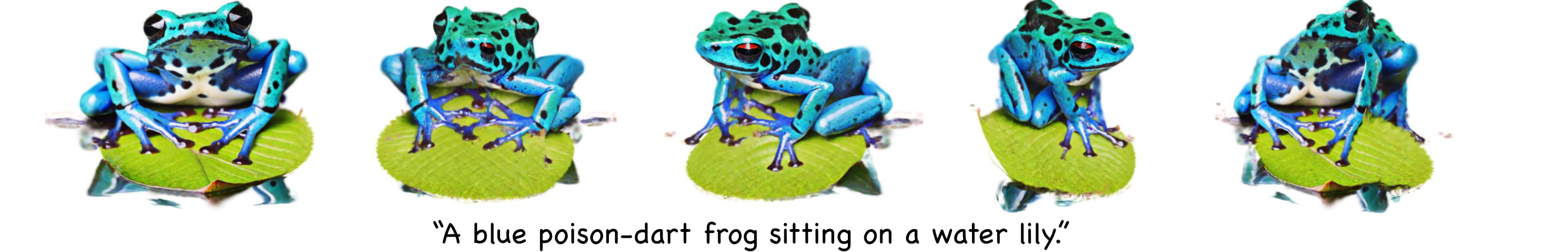"} 
    \caption{Failure case \uppercase\expandafter{\romannumeral1}. The 3D asset is affected by the Janus problem.}
    \label{fig:janus}
\end{figure}

\noindent\textbf{Failure case \uppercase\expandafter{\romannumeral1}: Janus Problem}. Despite proposing multiple strategies to improve the overall generation quality, our method still suffers from the Janus problem and occasionally generates multi-view inconsistent results (\Fref{fig:janus}).
A solution is to change the random seed or give a better initialization.

\begin{figure}[H]
    \centering
    \includegraphics[width=\textwidth]{"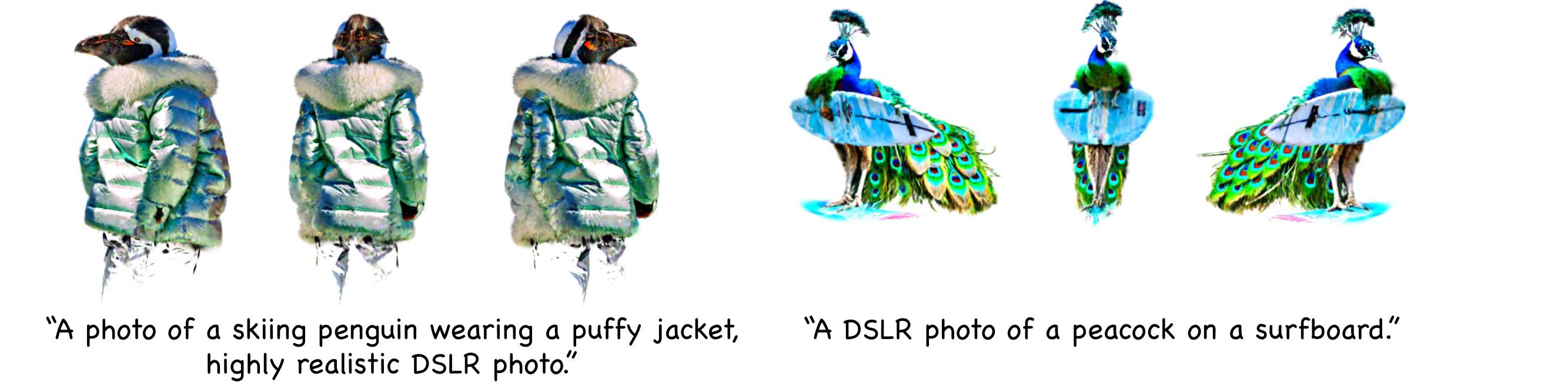"} 
    \caption{Failure case \uppercase\expandafter{\romannumeral2}. The 3D assets are generated based on a compositional prompt.}
    \label{fig:sup_fail}
\end{figure}

\noindent\textbf{Failure case \uppercase\expandafter{\romannumeral2}: Compositional Prompt}. Our method performs poorly in generating high-quality compositional objects (\Fref{fig:sup_fail}), possibly due to a bad initialization with insufficient Gaussian points to model different components or a lack of relevant techniques to combine those components correctly. We regard improving this as a future work.

\section{Example Question}
\label{sec:example}
We show an example question from our user study. We use Google Forms to collect the volunteers' responses.
\begin{figure}[H]
    \centering
    \includegraphics[width=0.9\textwidth]{"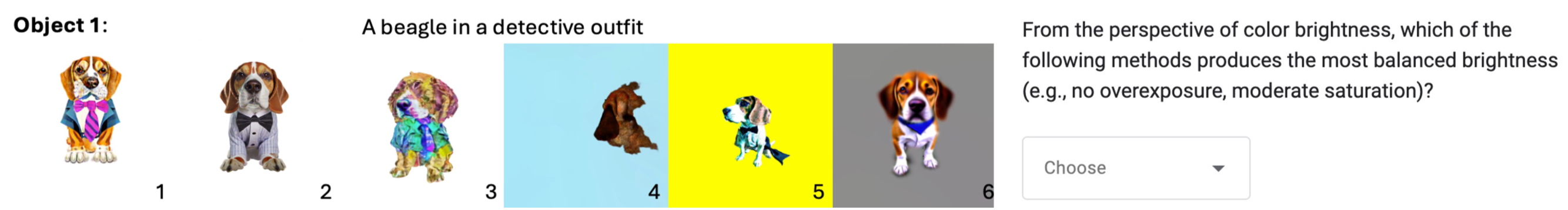"} 
    \caption{Example questions used in the user study.}
    \label{fig:sup_question}
\end{figure}

\section{Visualization of Accumulated Brightness}
\label{sec:acc}
\begin{figure}[H]
\includegraphics[width=0.9\textwidth]{"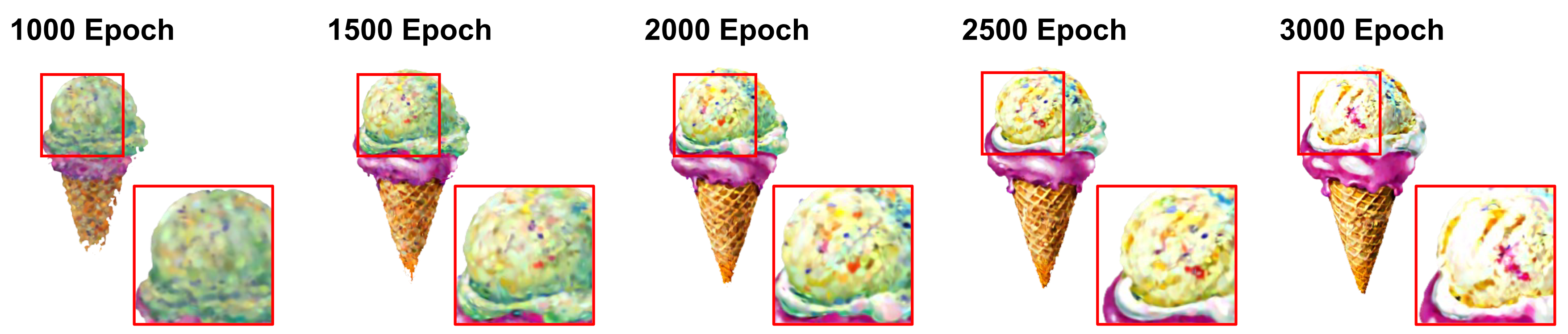"} 
    \caption{An illustration of brightness accumulation. The highlight point (red box) in early-stage training will be kept and spread to its surroundings, causing over-saturation.}
    \label{fig:highlight}
\end{figure}

We show an example of accumulated brightness during the generation process of the proposed method without BEG in \Fref{fig:highlight}.

\newpage
\section{Proof of Lemma 1}
\label{sec:proof}
\begin{lemma}[\cite{zheng2024trajectory, song2023consistency, kim2023consistency, wu2024consistent3d}]
\label{the:upcm2}
Let $\Delta t =\max \left\{\left|\delta_{k}\right|\right\}$, $k \in [0, ..., n_s]$, where $n_s$ is the index of $\delta$ at time step $s$, and $F_{\theta}(\cdot, \cdot)$ is the origin prediction function grounded on the empirical PF-ODE. Assume $F_{\theta}$ satisfies the Lipschitz condition, if there is a $\rvx_\pi$ satisfying $\mathcal{L}_{\rm{CC}}(\xi) = 0$, given an image $\rvx_0 \sim p_{\text{data}}(\rvx)$, for any $t, s, e\in [\![0,..., T]\!]$ with $t > s > e$,  we have:
{
\begin{equation}
\begin{split}
\mathop{\textnormal{sup}}\limits_{t, s, \rvx_\pi} \Vert \hat{\rvx}_{\{\pi, e\}}, \hat{\rvx}_{\{0,e\}} \Vert_2  = \mathcal{O} \left((\Delta t)^{p}\right) (T - e),
\end{split}
\end{equation}
}$\hat{\rvx}_{\{0,e\}}$ is the distribution of $\rvx$ diffused to time $e$, $p$ is the order of the ODE solver.
\begin{proof}
Given $\mathcal{L}_{\rm{CC}}(\xi) = 0$, for any $\pi$, $t, s$, and $e$, we have:
{
\begin{equation}
\label{eq:error_comp}
    G_\theta\left(\hat{\mathbf{x}}_{\{\pi, t\}} ; t, e, y\right)\equiv G_\theta\left(\hat{\mathbf{x}}_{\{\pi, s\}} ; s, e, y\right).
\end{equation}
}Defining $\hat{\mathbf{x}}_{\{\pi, t_k\}}=\mathbf{x}_{t_k}$ for simplicity,  
\Eref{eq:error_comp} can be rewritten as:
{
\begin{equation}
\label{eq:error_simp}
    G_\theta\left(\mathbf{x}_s ; s, e, y\right)\equiv G_\theta\left({\mathbf{x}}_t; t, e, y\right),
\end{equation}
}where, $\rvx_t$ is obtained by $\hat{\rvx}_s$ through DDIM inversion. For a more general expression, 
we represent distillation error~\cite{wu2024consistent3d} at $t_k>e$ as:
{
\begin{equation}
    e^{t_k}_e = G_\theta(\mathbf{x}_{t_k}; t_k, e, y) - \hat{\mathbf{x}}_{\{0, e\}}.
\end{equation}}It is straightforward that at the boundary timestep $e$, the error is,
\begin{equation}
    e^{e}_e = \hat{\mathbf{x}}_{\{\pi, e\}} - \hat{\mathbf{x}}_{\{0, e\}}.
\end{equation}We then derive $e^{t_k}_e$ at $t_k$ as:
{
\small
\begin{align}
\begin{aligned}
\boldsymbol{e}^{t_k}_e= & G_\theta\left(\mathbf{x}_{t_k}; t_k, e, y\right)-\hat{\mathbf{x}}_{\{0, e\}},\\
= & G_\theta\left(\bar{\mathbf{x}}_{\{t_k \rightarrow t_{k+1}\}}; t_{k+1},e, y\right)-\hat{\mathbf{x}}_{\{0, e\}}, \\
= & G_\theta\left(\bar{\mathbf{x}}_{\{t_k \rightarrow t_{k+1}\}}; t_{k+1},e, y\right)-G_\theta\left(\mathbf{x}_{t_{k+1}}; t_{k+1},e, y\right) +G_\theta\left(\mathbf{x}_{t_{k+1}}; t_{k+1},e, y\right)-\hat{\mathbf{x}}_{\{0, e\}}, \\
= & G_\theta\left(\bar{\mathbf{x}}_{\{t_k \rightarrow t_{k+1}\}}; t_{k+1},e, y\right)-G_\theta\left(\mathbf{x}_{t_{k+1}}; t_{k+1},e, y\right)+\boldsymbol{e}_{e}^{t_{k+1}}.
\end{aligned}
\end{align}
}According to Lipschitz condition on $G_\theta$, we derive:
{
\begin{align}
\begin{aligned}
\left\|\boldsymbol{e}^{t_{k}}_e\right\| & \leq\left\|G_\theta\left(\bar{\mathbf{x}}_{\{t_{k}\rightarrow t_{k+1}\}}; t_{k+1}, e, y\right)-G_\theta\left(\mathbf{x}_{t_{k+1}}; t_{k+1},e, y\right)\right\|+\left\|\boldsymbol{e}^{t_{k+1}}_e\right\|, \\
& \leq L\left\|\bar{\mathbf{x}}_{\{t_{k}\rightarrow t_{k+1}\}}-\mathbf{x}_{t_{k+1}}\right\|+\left\|\boldsymbol{e}^{t_{k+1}}_e\right\|, \\
& \stackrel{(i)}{=}\left\|\boldsymbol{e}^{t_{k+1}}_e\right\|+\mathcal{O}\left(\left(t_k-t_{k+1}\right)^{p+1}\right),
\end{aligned}
\end{align}
}where $(i)$ hold according to the local error of Euler solver~\cite{wu2024consistent3d, song2023consistency}. Iteratively, we can obtain the upper bound of distillation error at $e$:
{
\begin{align}
    \begin{aligned}
\left\|\boldsymbol{e}^e_e\right\|_2 & \leqslant \sum_{k=n_e}^{N-1} \mathcal{O}\left(\left(t_{k+1}-t_k\right)^{p+1}\right) + \|e^{T}_e\|,\\
& \stackrel{(ii)}{\approx} \sum_{k=n_e}^{N-1}\left(t_{k+1}-t_k\right) \mathcal{O}\left((\Delta t)^p\right),\\
& = \mathcal{O}\left((\Delta t)^p\right) \sum_{k=n_e}^{N-1}\left(t_{k+1}-t_k\right),\\
& =\mathcal{O}\left((\Delta t)^p\right)\left(T-e\right), \\
\end{aligned}
\end{align}
}where $N$ is the index of time $T$, $(ii)$ holds because $\|e^{T}_e\|$ describes the KL divergence between $\mathbf{x}_0$ and $G_\theta\left(\hat{\mathbf{x}}_{\{\pi, T\}} ; T, e, y\right)$, where $\hat{\mathbf{x}}_{\{\pi, T\}}$ follows the normal distribution. For a well-trained diffusion model, $\|e^{T}_e\|$ should be bounded and remain constant at $e$, which can be ignored. The proof is completed.
\end{proof}
\end{lemma}
According to Lemma.~\ref{the:upcm2}, we prove that the proposed CC loss will converge within a lower error bound than CDS loss in~\cite{wu2024consistent3d} when $e > 0$.


\newpage
\section{Additional Ablation Studies}
\label{sec:abl}
\begin{figure}[H]
    \centering
    \includegraphics[width=\textwidth]{"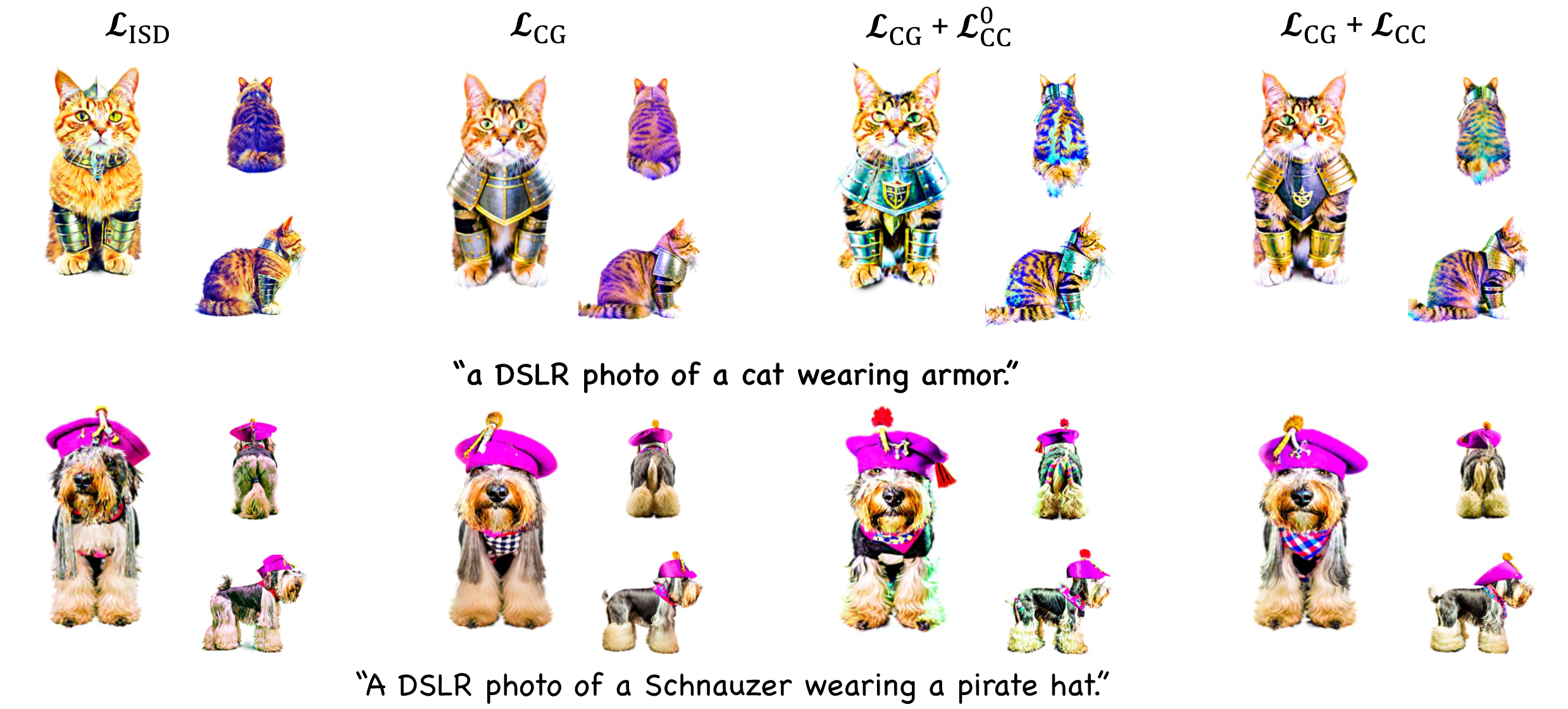"} 
    \caption{Qualitative comparison regarding different variances of $\mathcal{L}_{\text{CC}}$ and $\mathcal{L}_{\text{CG}}$. From left to right, 3D asset generated by $\mathcal{L}_{\text{ISD}}$, $\mathcal{L}_{\text{CG}}$, $\mathcal{L}_{\text{CG}} + \mathcal{L}_{\text{CC}}^0$, and $\mathcal{L}_{\text{CG}} + \mathcal{L}_{\text{CC}}$, respectively.}
    \label{fig:abl10}
\end{figure}

\textbf{Effect of CC and CG}. We conduct additional ablation studies to evaluate the effect of $\mathcal{L}_{\text{CC}}$ and $\mathcal{L}_{\text{CG}}$. Specifically, we compare $\mathcal{L}_{\text{CC}} + \mathcal{L}_{\text{CG}}$ with $\mathcal{L}_{\text{ISD}}$ (LucidDreamer~\cite{liang2023luciddreamer}), $\mathcal{L}_{\text{CG}}$, and $\mathcal{L}_{\text{CG}} + \mathcal{L}^0_{\text{CC}}$, where we note $\mathcal{L}^0_{\text{CC}}$ as $\mathcal{L}_{\text{CC}}$ with $e\equiv 0$. As shown in~\Fref{fig:abl10}, we find the proposed $\mathcal{L}_\text{CG}$ has higher generation quality compared to the LucidDreamer. $\mathcal{L}_{\text{CC}}$ will further add more details to the generated results in the generated 3D asset. Comparing with $\mathcal{L}_{\text{CG}}+\mathcal{L}^0_{\text{CC}}$ that shares the same distillation error bound with CDS~\cite{wu2024consistent3d}, we found the color distortion becomes more severe and tend to be over-exposed, validating the effect of proposed $\mathcal{L}_{\text{CC}}$ in reducing distillation error.

\noindent \textbf{Effect of High Order ODE-Solver}. We try 2nd order DPM-Solver at denoising steps, and the results indicate differences in illumination and details, as shown in~\Fref{fig:2nd}. 

{
\begin{figure}[H]
\floatbox[{\capbeside\thisfloatsetup{capbesideposition={right,top}}}]{figure}[\FBwidth]
{\caption{Generated views of GCS+BEG under a low CFG weight ($w=7.5$) by using first-order (right) and second-order DPM-Solver.}\label{fig:2nd}}
{\includegraphics[width=8cm]{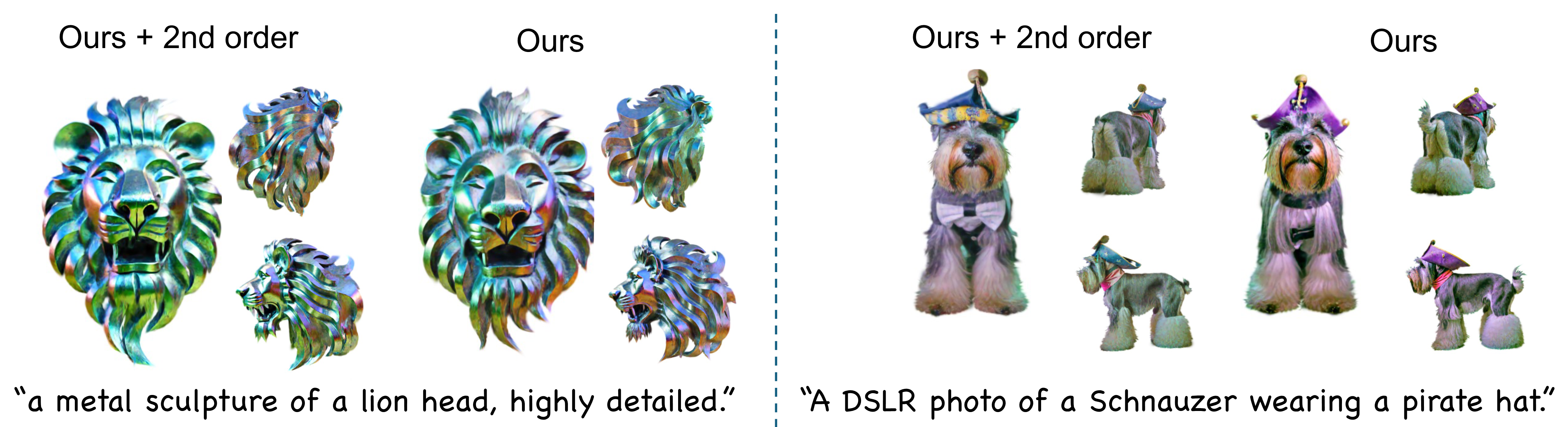}}
\end{figure}
}

\newpage
\section{Additional Qualitative Evaluation}
\begin{figure}[H]
    \centering
    \includegraphics[width=\textwidth]{"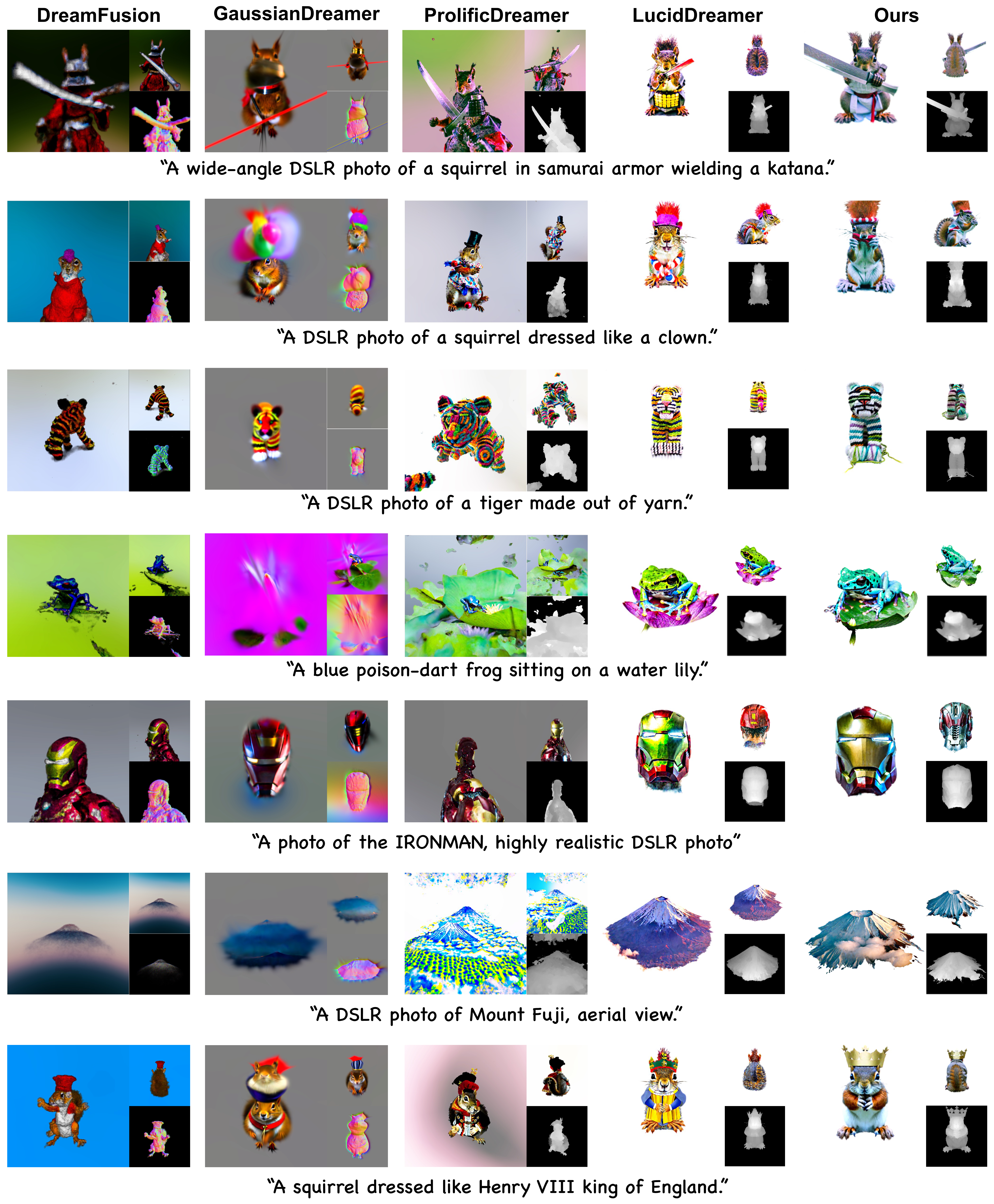"} 
    \caption{Additional qualitative comparison among DreamFusion~\cite{poole2022dreamfusion} (column 1), GaussianDreamer~\cite{yi2023gaussiandreamer} (column 2),  ProlificDreamer~\cite{wang2024prolificdreamer} (column 3), LucidDreamer~\cite{liang2023luciddreamer} (column 4), and our method (column 5).}
    \label{fig:sup_qual}
\end{figure}

In this section, we show more quality comparison between the proposed GCS,  DreamFusion~\cite{poole2022dreamfusion}, GaussianDreamer~\cite{yi2023gaussiandreamer}, ProlificDreamer~\cite{wang2024prolificdreamer}, and LucidDreamer~\cite{liang2023luciddreamer} in \Fref{fig:sup_qual}.
\label{sec:qe}

\end{document}